\documentclass[lettersize,journal]{IEEEtran}
\usepackage{amsmath,amsfonts}
\usepackage{algorithmic}
\usepackage{algorithm}
\usepackage{array}
\usepackage[caption=false,font=normalsize,labelfont=sf,textfont=sf]{subfig}
\usepackage{textcomp}
\usepackage{stfloats}
\usepackage{url}
\usepackage{verbatim}
\usepackage{graphicx}
\usepackage{cite}
\usepackage{graphicx}
\usepackage{multirow}
\usepackage{booktabs}
\usepackage{amsmath,amssymb} 

\begin{document}

\title{Cross-Modal Graph with Meta Concepts for Video Captioning}

\author{Hao~Wang,
        Guosheng~Lin,
        Steven~C.~H.~Hoi,~\IEEEmembership{Fellow,~IEEE}
        and~Chunyan~Miao
\IEEEcompsocitemizethanks{\IEEEcompsocthanksitem Hao Wang, Guosheng Lin and Chunyan Miao are with School of Computer Science and Engineering, Nanyang Technological University. E-mail: \{hao005,gslin,ascymiao\}@ntu.edu.sg.
\IEEEcompsocthanksitem Steven C. H. Hoi is with Singapore Management University and Salesforce Research Asia. E-mail: chhoi@smu.edu.sg.
}
\thanks{Corresponding authors: Chunyan Miao and Guosheng Lin.}
}

\markboth{Journal of \LaTeX\ Class Files,~Vol.~14, No.~8, August~2021}%
{Shell \MakeLowercase{\textit{et al.}}: A Sample Article Using IEEEtran.cls for IEEE Journals}

\maketitle

\begin{abstract}
Video captioning targets interpreting the complex visual contents as text descriptions, which requires the model to fully understand video scenes including objects and their interactions. Prevailing methods adopt off-the-shelf object detection networks to give object proposals and use the attention mechanism to model the relations between objects. They often miss some undefined semantic concepts of the pretrained model and fail to identify exact predicate relationships between objects. In this paper, we investigate an open research task of generating text descriptions for the given videos, and propose Cross-Modal Graph (CMG) with meta concepts for video captioning. Specifically, to cover the useful semantic concepts in video captions, we weakly learn the corresponding visual regions for text descriptions, where the associated visual regions and textual words are named cross-modal meta concepts. We further build meta concept graphs dynamically with the learned cross-modal meta concepts. We also construct holistic video-level and local frame-level video graphs with the predicted predicates to model video sequence structures. We validate the efficacy of our proposed techniques with extensive experiments and achieve state-of-the-art results on two public datasets. 
\end{abstract}

\begin{IEEEkeywords}
Video Captioning, Vision-and-Language.
\end{IEEEkeywords}

\section{Introduction} \label{sec:introduction}
\IEEEPARstart{V}{ideo} captioning aims to give precise descriptions for input videos, which benefits many relevant applications such as human-computer interaction and video retrieval \cite{yu2017end}. Although this task is trivial for humans, it can be very challenging for machine to achieve satisfying results. The main reasons are (i) videos contain complex spatial and temporal information within changing scenes, and (ii) captions have underlying syntax, including objects and their relationships. That means the captioning model is required to interpret input videos by identifying multiple objects as well as predicting their interactions.

\begin{figure}
\begin{center}
\includegraphics[width=0.4\textwidth]{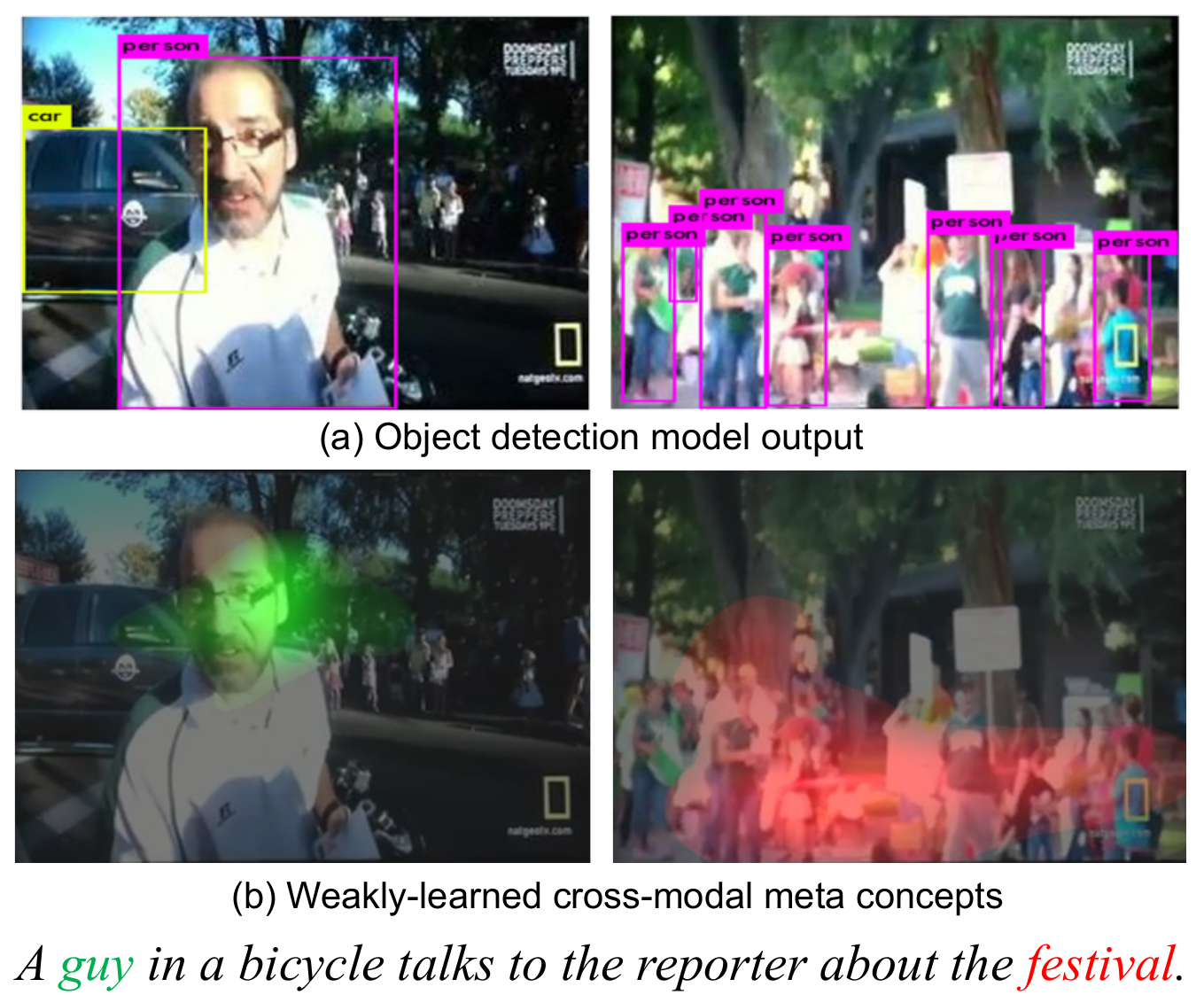}
\end{center}
   \caption{\textbf{Two video frames extracted from MSR-VTT dataset as well as the caption.} We show the differences between object detection model outputs and our weakly-learned cross-modal meta concepts. In the top row, object proposals are produced with YOLOv3 model \cite{redmon2018yolov3}, which fails to detect the undefined semantic concept of pretrained model: \emph{festival}. In the bottom row, we show the predicted visual regions for the given captioning words with our proposed model, where the green region and red region refer to \emph{guy} and \emph{festival} in the caption respectively.}
\label{fig:demo}
\end{figure}

Prior works \cite{wang2018m3,jin2016describing,ramanishka2016multimodal} adopt 3D convolutional models to incorporate motion and temporal feature representations, which consider global visual information for videos but lack object-level representations. Recently, some works \cite{pan2020spatio, zhang2020object, zheng2020syntax} focus on exploiting more fine-grained video representations by building graphs based on object proposals that are given by pretrained model. However, there are two main limitations of their work affecting model performance: (i) the pretrained object detector may fail to detect the undefined semantic concepts (Figure \ref{fig:demo}) in video frames and (ii) the predicate relationships between objects are not explicitly predicted.

In terms of the first limitation, as is shown in Figure \ref{fig:demo}, when we adopt a pretrained detection model to predict possible existing objects in the video, since the model is not trained on the video captioning datasets, it would miss some semantic concepts that are not defined during pretraining. For instance, in the top row of Figure \ref{fig:demo}, YOLOv3 model \cite{redmon2018yolov3} fails to detect \emph{festival}. Besides, there are many animation clips in given video captioning datasets \cite{xu2016msr,chen2011collecting}, the pretrained detection model may also fail at these scenarios. As a result, the constructed graphs can hardly provide enough fine-grained semantic information for the caption generation process.
Regarding the second limitation above, previous works \cite{pan2020spatio, zhang2020object} only construct soft connections between detected objects, which is realized by attention mechanism or similarity computation. This would result in the computed relationships between objects being implicit, in other words, the predicate information remains unclear. However, predicates play an important role in caption generation \cite{wang2019controllable, zheng2020syntax}, which can guide the model to be aware of the syntax and the exact interactions between objects. 

To address the aforementioned limitations, we propose the Cross-Modal Graph (CMG) with meta concepts for video captioning. Specifically, to cover missed semantic concepts by the pretrained detection model, we propose to learn the \emph{cross-modal meta concepts}, consisting of the visual and semantic meta concepts, which are defined as the visual regions and their corresponding semantic words in captions. Since we do not have explicit pixel labeling, we adopt a weakly-supervised approach to uncover attended visual regions for words, which are learned through training the decoder to generate video captions.
We further use the learned meta concepts as pseudo masks to train a localization model, which is incorporated into our captioning model and predicts the underlying meta concepts when processing video keyframes. After obtaining multiple predicted meta concepts, we construct meta concept graphs dynamically to output representations. In the bottom row of Figure \ref{fig:demo}, we show the learned cross-modal meta concepts, which can find the attended regions with their corresponding semantic classes.
To give explicit predicate relationships between objects in videos, we also use a scene graph detection model \cite{tang2020sggcode} to predict object pairs along with their predicates. Based on the detected scene graphs, we then build holistic video-level and local frame-level graphs to give multi-scaled video structure representations.

Our contributions can be summarized as:
\begin{itemize}
   \item We propose to use a weakly-supervised learning method to discover the corresponding visual regions of the given words of target captions, i.e. cross-modal meta concepts, which are used as the pseudo masks to train a localization model. This localization model is applied to predict meta concepts for the caption generation model.
   \item We build three types of cross-modal graph representations, i.e. the dynamically constructed meta concept graphs, holistic video-level and local frame-level video graphs from the detected scene graphs.
   \item We conduct extensive experiments to verify the usefulness of various modules of our model, and our model achieves state-of-the-art results on MSR-VTT \cite{xu2016msr} and MSVD \cite{chen2011collecting} datasets.
\end{itemize}

\section{Related Work}

\subsection{Video Understanding}
Most video captioning works \cite{yao2015describing,wang2018m3,xu2017learning,wang2018reconstruction,jin2016describing,pan2017video,gan2017stylenet,long2018video,yi2019clevrer,chen2021grounding} are built with encoder-decoder architecture, where a CNN is adopted to extract video features and an RNN is used to recurrently generate the descriptions. Earlier work improves input video representations with different manners. Xu et al. \cite{xu2017learning} incorporate multimodal attention over LSTM to obtain video representations. Wang et al. \cite{wang2019learning} exploit a cycle-consistent idea to reproduce the visual contents after caption generation. In \cite{chen2018less}, Chen et al. introduce a frame picking module to select video keyframes. Recent video captioning works \cite{wang2019controllable,zheng2020syntax,pan2020spatio,zhang2020object} focus on improving correspondence between videos and captions. Both Zhang et al. \cite{zhang2019object} and Zheng et al. \cite{zheng2020syntax} propose to use Part-of-Speech (POS) information to guide video captioning process, where they generate objects first and consequently output actions of captions. Modeling interactions between objects in videos \cite{pan2020spatio,zhang2020object} is an emerging way for video captioning. Specifically, Pan et al. \cite{pan2020spatio} propose to construct a spatial-temporal graph and use knowledge distillation way to integrate the object features with video visual features. Zhang et al. \cite{zhang2020object} take an external large-scale language model to boost the original language decoder learning. Duan et al. \cite{duan2018weakly} propose to use the weakly supervised learning method to address the video captioning problem. RCG \cite{zhang2021open} introduces to train an additional retrieval model for the video captioning task, Zhang et al. \cite{zhang2021open} firstly train an individual video-to-text retrieval model, where the corresponding descriptive sentences can be retrieved based on the given videos. During the caption generation phase, RCG learns a weighted sum between the retrieval words distribution and the captioning decoder output vocabulary distribution, based on which the final captions can be generated. Video reasoning \cite{wu2021star,ding2021dynamic,chen2022comphy} enables the model to build the relationships between visual concepts, which can infer objects and their interactions from videos and language. Using video reasoning can also benefit some down-steam tasks \cite{wu2021star,ding2021dynamic,chen2022comphy}.

To construct better video representations, Wei et al. \cite{wei2019neural} explicitly separate the consistent features and the complementary features from the mixed information and harness their combination, aiming to improve the complementarity of different modality information in video. To reduce the computational overhead, Zheng et al. \cite{zheng2020dynamic} design a dynamic sampling module to improve the discriminative power of learned clip-level classifiers and increase the inference efficiency during testing. SibNet \cite{liu2020sibnet} proposes a novel framework to extract video temporal features better, while it fails to consider the video-caption correspondence and did not use the audio features. Therefore, SibNet has inferior results than our proposed method. SwinBERT \cite{lin2021swinbert} adopts an end-to-end architecture, where the video features are extracted online. Technically, the video tokens and word tokens are simultaneously fed into the model, hence the attention masks can be learned, which reflect the visual semantic correspondence. However, end-to-end training requires intensive computation resources. By contrast, our proposed method is more lightweight.

\begin{figure*}[htbp!]
\begin{center}
\includegraphics[width=0.8\textwidth]{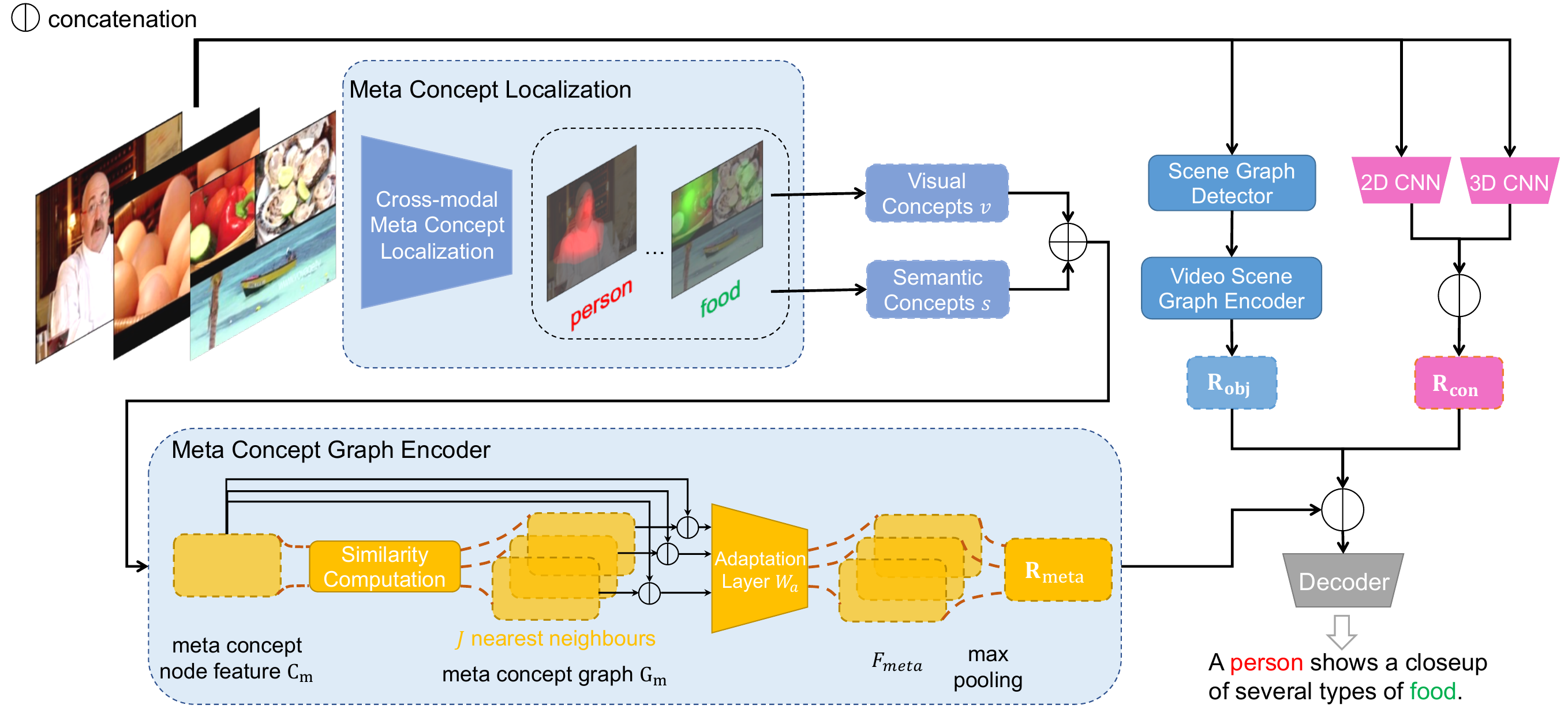}
\end{center}
  \caption{\textbf{CMG: Cross-Modal Graph with meta concepts:} Our proposed framework to achieve effective video captioning, which consists of three modules: meta concept localization model, meta concept graph encoder and video scene graph encoder. In the first module, we first use a weakly-supervised method to learn cross-modal meta concepts for the given datasets, which are then used as the pseudo masks to train a localization model to output the visual regions and semantic information correlated with the target captions. In the meta concept graph encoder, to enable adaptive information flow across the video, we encode the predicted meta concepts $\mathrm{C_m}$ with a dynamic graph construction way $\mathrm{G_{m}}$ and give meta concept representation $\mathbf{R_{meta}}$. In the video scene graph encoder, we take a pretrained model to detect scene graphs for video frames, and then build local frame-level and holistic video-level graphs $\{\mathrm{G_{f}}, \mathrm{G_{v}}\}$ to give video object structural representation $\mathbf{R_{obj}}$. Finally, we concatenate video context features $\mathbf{R_{con}}$, $\mathbf{R_{meta}}$ and $\mathbf{R_{obj}}$ as decoder inputs to generate video captions.}
\label{fig:framework}
\end{figure*}

\subsection{Concept Prediction}
Learning semantic concepts from visual input has been validated to be useful in the captioning task \cite{you2016image,gan2017semantic,wu2016value}, where they mainly use a multi-label classification to predict the hidden high-level concepts. To be specific, You et al. \cite{you2016image} predict the semantic attributes from the given images before the captioning process, which are fused with the recurrent neural networks and the attention mechanism. Gan et al. \cite{gan2017semantic} propose a semantic compositional network for the image captioning task, where they detect the semantic concepts from the input images and the probability of each of them is adopted to compose the parameters of the caption generator. Wu et al. \cite{wu2016value} use the multi-label classification model to produce the predicted attributes, which can improve the performance of both the image captioning and VQA tasks on several benchmark datasets.

Recently, Zhou et al. \cite{zhou2019grounded} propose to use the grounding approach to find the attended visual regions for words in the given captions, which improves previous methods by introducing the localized visual representations. However, grounding methods \cite{zhou2019grounded,ma2020learning} are based on the object proposals that are produced by pretrained object detectors. That means these models can only find some object classes that are pre-defined. In contrast, our method is not constrained by the pretrained detector, we can predict a wider range of semantic concepts and the corresponded visual regions.

VinVL \cite{zhang2021vinvl} and our proposed method are concurrent works. Specifically, VinVL introduced powerful detectors that can identify more than a thousand concepts including objects, relationships and attributes. However, their prediction ranges are still limited on the pretrained datasets. Given the MSR-VTT dataset contains a large portion of animations, the pretrained models may fail in these scenarios, while our proposed method gives reasonable outputs. Moreover, VinVL requires large-scale training data with annotations, and needs many computation resources for training. By contrast, we use the weakly-supervised method for training, and the training time is much less than VinVL. We provide a solution to generate semantic concepts when ones do not have large-scale labeled training data or enough computation resources.

Specifically, instead of using object proposals detected by the pretrained models as the graph nodes \cite{pan2020spatio,zhang2020object}, we adopt a weakly-supervised learning approach to localize the visual regions, which are aligned with the corresponding caption semantic concepts, and use them as the nodes of our proposed meta concept graph. In this way, our method is not limited by the pretrained detector, we can predict a wider range of meta concepts.

\subsection{Graph Models}
To model non-Euclidean structures such as graphs and trees, Graph Convolution Networks (GCN) \cite{kipf2017semi} are proposed to give graph structure representations. Later, graph attention networks (GAT) \cite{velickovic2018graph} introduce attention mechanism when encoding node features. In \cite{li2019deepgcns}, Li et al. further explore the effect of stacking deeper layers for GCN. However, both GCN and GAT require the pre-defined edge information as the input to compute the embedded graph features. To alleviate this limitation, Wang et al. \cite{wang2019dynamic} compute the node relationships and aggregate neighbourhood features at each iteration, so that we can compute the graph representations without the pre-defined edges and build the relationships between nodes dynamically.

There have been some efforts to apply graph models on the captioning task \cite{yang2019auto,li2019know,yao2018exploring}. To be specific, Yang et al. \cite{yang2019auto} use a pretrained detector to give predicted scene graphs of the images first, and then adopt a GCN to embed the scene graphs. In \cite{yao2018exploring}, semantic graph is built with directional edges on the detected object regions, where they also use GCN to embed and produce the contextual features.
In terms of video related tasks, spatial and temporal information is of great significance. Xu et al. \cite{xu2020g} perform graph convolution on segmented video snippets to leverage both spatial and temporal context for action localization. Liu et al. \cite{liu2019learning} take videos to be 3D point clouds in the spatial-temporal space. Wang et al. \cite{wang2018videos} build video graphs from the computed similarity and use GCN model to give representations for video action recognition. Zhang et al. \cite{zhang2021multi} propose to tackle the problem of temporal language localization in videos with the multi-modal interaction graph convolutional network.

In this paper, we construct various types of graphs with different graph embeddings, where we use dynamic graph embedding way to model the cross-modal meta concepts across the whole video, and include the sequence information to the detected scene graphs in video frames. We also give a detailed analysis of their effects on video captioning tasks.

\begin{figure*}
\begin{center}
\includegraphics[width=0.8\textwidth]{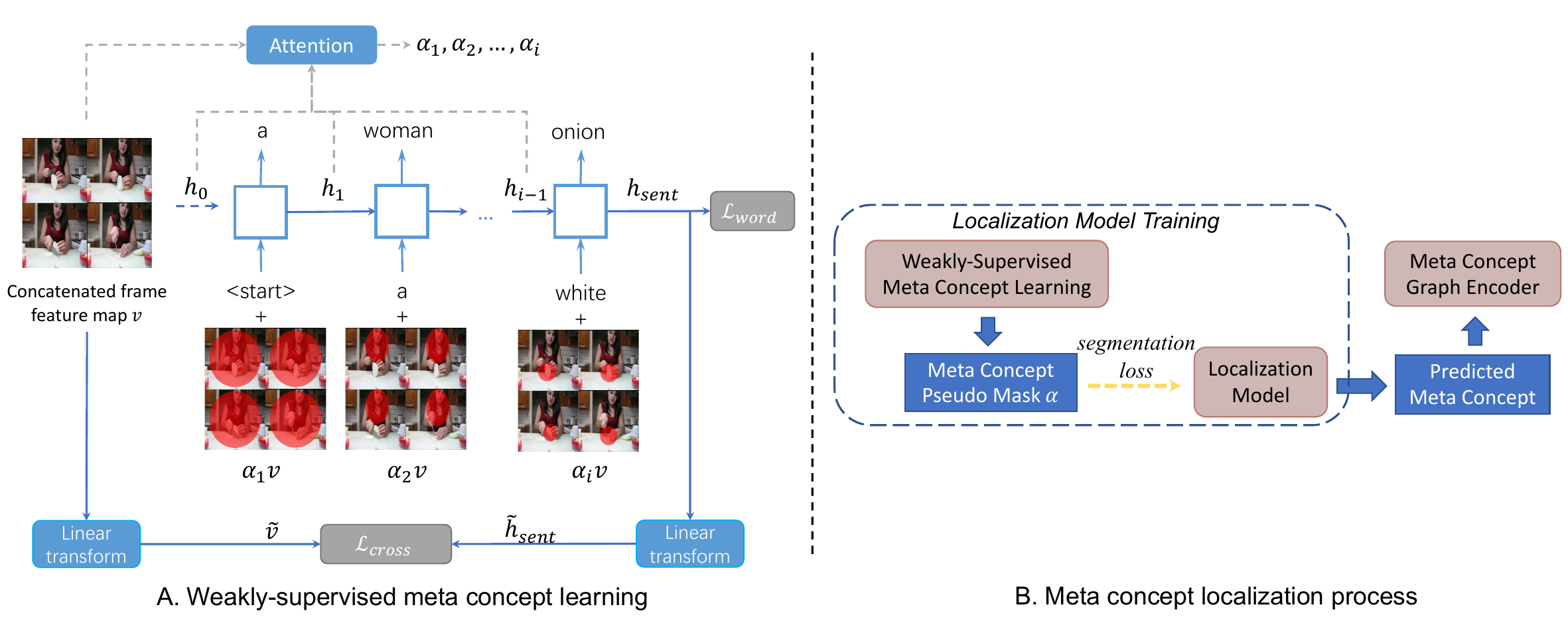}
\end{center}
  \caption{\textbf{The demonstration of our proposed weakly-supervised meta concept learning process.} \textbf{Step A:} We use an LSTM model to localize the visual meta concepts, where the input is the concatenated feature maps $v$ from ResNet-101. In each training step, we compute the attention map $\alpha_i$ from $h_{i-1}$ and $v$, then use $\alpha_i v$ to generate each word, which is supervised by cross-entropy loss $\mathcal{L}_{word}$. We also give cross-modal alignments over visual and sentence features by $\mathcal{L}_{cross}$ to improve the correspondence between vision and language. \textbf{Step B:} The learned attention maps $\alpha$ of each caption word from step A are used as the pseudo masks, to train a localization model. The localization model takes the video frames as the inputs, and predicts the visual regions (visual meta concepts) along with the corresponding words (semantic meta concepts). The predicted cross-modal meta concepts are fed into the meta concept graph encoder to give the graph features.
  }
\label{fig:learnmeta}
\end{figure*}

\section{Method}
The proposed Cross-Modal Graph (CMG) with meta concepts for video captioning is presented in Figure \ref{fig:framework}. Here we define \emph{cross-modal meta concepts} as: the visual regions and the corresponding caption words, which cover both visual and semantic information. Note that for computational efficiency, we first find the absolute difference between frames, then we sort the computed difference and pick the top $N$ frames to be keyframes, which are inputs for our proposed model. Our goals have two folds: (i) learn informative cross-modal meta concepts and bridge the gap between videos and their descriptions and (ii) capture the complex relationships between various objects in the video. To this end, we design a novel model that consists of three modules: meta concept localization model, meta concept graph encoder and video scene graph encoder.

In the meta concept localization model, we use a weakly-supervised manner to discover the cross-modal meta concepts with caption guidance, which is shown in Figure \ref{fig:learnmeta}. It is observed that there are multiple text descriptions for each video, we use a sentence scene graph parsing tool\footnote{https://github.com/vacancy/SceneGraphParser} to extract object tokens $\mathrm{T}=\{t_1, \dots, t_k\}$ from all captions, then we group $\mathrm{T}$ based on synonym rules and sort them by frequency order, we take the top $K$ groups of synonyms $\mathrm{T'} \in \mathrm{T}$ to be semantic classes. We adopt the attended visual regions for $\mathrm{T'}$ to be pseudo masks to train a semantic segmentation network, which is used to localize objects from $\mathrm{T'}$ during caption generation. 
Note that the localization model is trained individually to ease training difficulty, otherwise we need to jointly train the whole model and the training speed would be slow. 

With the trained localization model, we are able to localize caption word $t_i \in \mathrm{T'}$ in video frames and get visual representations $v_i$ for the attended regions. We encode the one-hot vectors of $t_i$ as semantic information, denoting as $s_i$. Our predicted meta concepts can be formulated as: $\mathrm{C_m} = \{c_1, \dots, c_L | c_i=[v_i, s_i]\}_{i=1}^L$, where $L$ is the number of predicted meta concepts across the video frames. The cross-modal meta concept representation $c_i$ is the sum of $v_i$ and $s_i$. Since $\mathrm{C_m}$ have implicit interactions between each other, we propose to use a dynamic graph construction way to give output representation $\mathbf{R_{meta}}$ of meta concept graph encoder.

\begin{figure*}
\begin{center}
\includegraphics[width=0.9\textwidth]{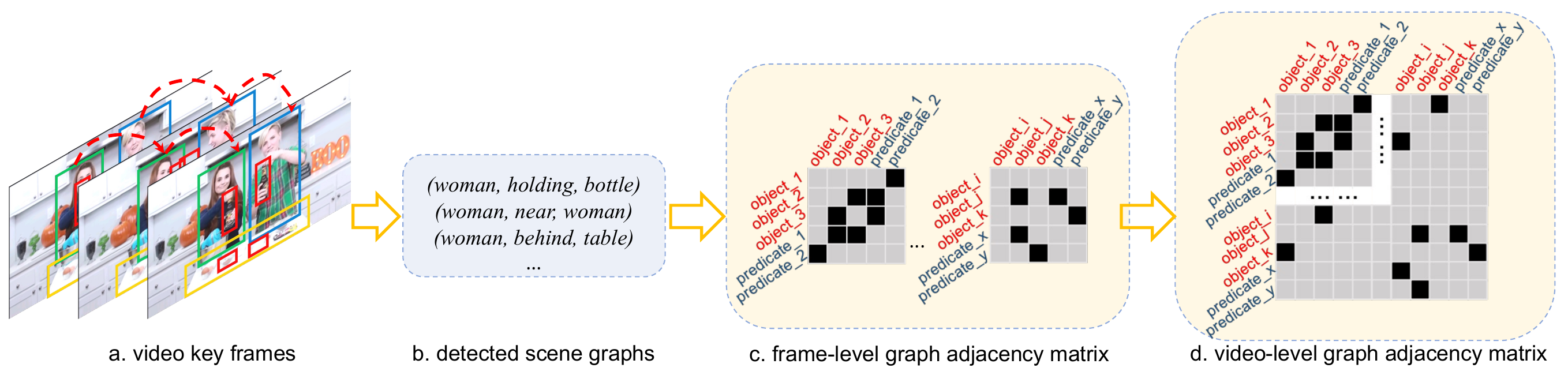}
\end{center}
  \caption{\textbf{The demonstration of our constructed frame-level and video-level graphs.} We use the pretrained scene graph detector to generate the detected scene graphs for the given video keyframes. For each scene graph triplet, we construct two edges between the objects and the predicates. There are multiple frame-level scene graphs in videos, while we only plot two of them for simplicity. The video-level graph is built by grouping frame-level graphs, where we connect nodes from the adjacent frames with high similarity. The black blocks denote connections between nodes.}
\label{fig:graph}
\end{figure*}

In the video scene graph encoder module, we adopt the method of \cite{tang2020unbiased} to generate the scene graphs for video frames. Technically, Tang et al. \cite{tang2020unbiased} propose to build a causal graph for scene graph generation. They firstly perform traditional scene graph training. In the second step, the counterfactual causality can be drawn from the trained graph, where they further infer the effect from the bad bias and attempt to remove it. 
With the generated scene graphs on video frames, we take two ways to build the object graph $\mathrm{G_{obj}} = \{\mathrm{G_{f}}, \mathrm{G_{v}}\}$, i.e. 
local frame-level graphs $\mathrm{G_{f}}$ and holistic video-level graphs $\mathrm{G_{v}}$, as shown in Figure \ref{fig:graph}. In $\mathrm{G_{f}}$, we first encode scene graphs through GAT \cite{velickovic2018graph} and obtain graph feature for each frame. We then adopt a transformer \cite{vaswani2017attention} to embed frame scene graph features to model sequence dependency between frames, giving $\mathbf{R}_{\mathrm{G_{f}}}$. We follow \cite{zhang2020object,pan2020spatio} to construct $\mathrm{G_{v}}$, which is based on the cosine similarity and interaction over union (IoU) between adjacent frame objects. We also use GAT to encode $\mathrm{G_{v}}$ and output $\mathbf{R}_{\mathrm{G}_{v}}$. We concatenate $\mathbf{R}_{\mathrm{G_{f}}}$ and $\mathbf{R}_{\mathrm{G_{v}}}$, and get the video scene graph representations as $\mathbf{R_{obj}} = [\mathbf{R}_{\mathrm{G_{f}}}, \mathbf{R}_{\mathrm{G_{v}}}]$.

Apart from the aforementioned modules, we also follow \cite{cst_phan2017,zheng2020syntax} to do prepossessing and extract video context representations $\mathbf{R_{con}}$ given input videos. We concatenate all obtained features together and get $\mathbf{R_{dec}}= [\mathbf{R_{con}}, \mathbf{R_{meta}}, \mathbf{R_{obj}}]$ for the decoder to recurrently generate captions, where the decoder is a one-layer plain LSTM. We train our model by two methods: minimizing the cross-entropy loss or maximizing a reinforcement learning (RL) \cite{luo2018discriminability} based reward. 

\subsection{Weakly-Supervised Meta Concept Learning}
The learning process is shown in Figure \ref{fig:learnmeta}. We first learn the cross-modal meta concepts in a weakly-supervised approach, which are adopted as pseudo masks to train the localization model. Then we use the trained localization model to predict the meta concepts, which are embedded with meta concept graph encoder and further used to generate video captions. In other words, the meta concept model is trained to cover semantic concepts in captions. 

Different from using the detected bounding boxes of the pretrained model, we are able to customize the classes of cross-modal meta concepts for various datasets. To this end, we adapt the model of \cite{xu2015show} to video captioning datasets. Specifically, we randomly sample $4$ frames from picked keyframes for each video and use ResNet-101 \cite{he2016deep} to embed sampled frames to get the output feature maps from the last convolutional layer. We then concatenate $4$ extracted feature maps to be the video representation $v$ for model training. We use a LSTM to weakly learn the corresponding regions for words in captions, where we impose word-level constraints and cross-modal alignment. In the word-level training, we feed in the initialized hidden states $h_0$ and previous word $t_{i-1}$ for LSTM model:
\begin{equation}
\begin{aligned}
    \alpha_i = softmax(W_{f}  ReLU(W_v v + W_h  h_{i-1})), 
\end{aligned}
\end{equation}
\begin{equation}
    h_i = \textbf{LSTM} ( [W_e t_{i-1}, \alpha_i v], h_{i-1} ),
\end{equation}
where $[\cdot, \cdot]$ indicates concatenation operation, $W_v, W_h, W_{f}$ and $W_e$ denote different learnable embedding matrix and $\alpha_i$ corresponds to the attention weights on visual feature map $v$ for the $i$th word. That means we use the frame features with attention weights to be the context vector for LSTM. The cross-entropy loss is used to decode caption sequence for word-level training:
\begin{equation}
    p_i = softmax(W_p h_i),
\end{equation}
\begin{equation}
    \mathcal{L}_{word} = -\sum_i \text{log } p(\hat{t}_i= t_i),
\end{equation}
where $p_i$ denotes the predicted probability over the vocabulary and $\hat{t}_i$ is the predicted word in training phase.

To improve the matching between video frame visual representations and caption text representations, we propose to use cross-modal alignment. Technically, we extract the final LSTM output to be sentence embedding $h_{sent}$, and apply linear transforms on $v$ and $h_{sent}$ to map them to the same dimension as $\widetilde{v}$ and $\widetilde{h}_{sent}$. We aim to impose alignment constraints over $\widetilde{v}$ and $\widetilde{h}_{sent}$, hence in the mini-batch, we sample $M$ video-caption pairs $\{(\widetilde{v}^i, \widetilde{h}_{sent}^i)\}_{i=1}^M$ from different identities, where $M$ is the batch size. We define only $\widetilde{v}^i$ matches $\widetilde{h}_{sent}^i$ while the rest $M-1$ $\widetilde{h}_{sent}$ are all mismatched with $\widetilde{v}^i$. We adopt triplet loss to achieve the alignment, the objective is given as:
\begin{equation}
\begin{aligned}
\mathcal{L}_{cross} =  & \sum_v\left[s(\widetilde{v}^{a}, \widetilde{h}_{sent}^{p})-s(\widetilde{v}^{a},\widetilde{h}_{sent}^{n})+m\right]_+ \; \\
& + \sum_{h_{sent}}\left[s(\widetilde{h}_{sent}^{a}, \widetilde{v}^{p})-s(\widetilde{h}_{sent}^{a}, \widetilde{v}^{n})+m\right]_+,
\end{aligned}
\end{equation}
where $s(\cdot, \cdot)$ denotes the Euclidean distance measurement, superscripts $a,p$ and $n$ refer to anchor, positive and negative instances respectively, and $m$ is the margin of error.

The whole training objective for weakly-supervised meta concept learning is given as:
\begin{equation}
    \mathcal{L}_{meta} = \mathcal{L}_{word} + \lambda \mathcal{L}_{cross},
\end{equation}
where $\lambda$ is the trade-off parameter.

To localize the corresponding meta concepts, we take $\alpha_i$ as pseudo masks to train another semantic segmentation model \cite{zhao2017pyramid} to infer meta concepts when generating video captions. We first extract all semantic concepts $\mathrm{T}$ from given captions and group them based on synonym rules, then the top $K$ classes of cross-modal meta concepts $\mathrm{T'} \in \mathrm{T}$ are taken to be the pseudo labels for segmentation model training. Since each video may have various meta concepts, we train the segmentation model with multi-label loss, i.e. the probability of each class is computed separately and optimized with a binary cross-entropy loss.

\subsection{Meta Concept Graph Encoder}
With the trained localization model, we can obtain a set of cross-modal meta concepts for given video: $\mathrm{C_m} = \{c_1, \dots, c_L | c_i=[v_i, s_i]\}_{i=1}^L$, where $L$ is the number of predicted meta concepts, $v_i$ and $s_i$ denote visual and semantic features from localization model output. We use the sum of $v_i$ and $s_i$ to be the representation of $c_i$.  Note the span of $\mathrm{C_m}$ is not restricted in a single frame, but covers all detected meta concepts from the picked keyframes. We propose to construct a dynamic graph $\mathrm{G_m} = \{\mathrm{C_m}, \mathrm{E_m}\}$ to integrate features of $\mathrm{C_m}$, where $\mathrm{C_m}$ is regraded as the node set and $\mathrm{E_m}$ denotes edge set.

Our target of constructing this dynamic graph is to capture interactions between semantically similar meta concepts, which can be defined by feature distance \cite{xu2020g}. Hence we use the k-nearest neighbour algorithm to build edges between nodes as follows:
\begin{equation}
    \mathrm{E_m} = \{(c_i, n_j^{(c_i)}) | j \in \{1, \dots, J\}\}_{i=1}^L,
\end{equation}
where $n_j^{(c_i)}$ denotes the $j$th neighbour of $c_i$. After we build edges for $\mathrm{C_m}$, we can get the adjacency matrix $\mathrm{A_m}$ for the constructed graph. Since we connect nodes dynamically during training phase, the model can keep updating node features and aggregating both intra- and inter-frame information.

Based on the obtained adjacency matrix $\mathrm{A_m}$ and node features $\mathrm{C_m}$, we perform graph convolution as follows:
\begin{equation}
    F_{meta}(\mathrm{C_m}, \mathrm{A_m}) = ([\mathrm{C_m^T}, \mathrm{A_m}\mathrm{C_m^T}]W_a)^T,
\end{equation}
where $[\cdot, \cdot]$ denotes concatenation operation, $W_a$ is a learnable adaption layer.
\begin{equation}
    \mathbf{R_{meta}} = \text{maxpool}(F_{meta}(\mathrm{C_m}, \mathrm{A_m})),
\end{equation}
We apply a max-pooling operation on the graph convolution output $F_{meta}(\mathrm{C_m}, \mathrm{A_m})$ and obtain the cross-modal meta concept graph representation $\mathbf{R_{meta}}$.

\subsection{Video Scene Graph Encoder}
In this module, we use an off-the-shelf scene graph model \cite{tang2020sggcode,tang2020unbiased}, to give video frame-level graph results $\mathrm{G_{f}}=\{\mathrm{G_{f}^1},\dots,\mathrm{G_{f}^N} | \mathrm{G_{f}^i} =(o_x, r_{xy}, o_y)^P\}_{i=1}^N$, where $P$ denotes the number of predicted relationship triplets, $N$ is the frame number and $o \rightarrow \mathbb{O}^{150}$, $r \rightarrow \mathbb{R}^{50}$, meaning we have $150$ classes of objects and $50$ types of predicate relationships. 

We build edges for $(o_x, r_{xy})$ and $(r_{xy}, o_y)$ respectively, and then we can obtain the adjacency matrix $\mathrm{A_{f}}$ for $\mathrm{G_{f}}$. We further extract node features for $\mathrm{G_{f}}$, where we take the one-hot vectors for each node and use a linear layer to encode them, then we are able to get the node features $\mathrm{F_{n}}$. $\mathrm{A_{f}}$ and $\mathrm{F_{n}}$ are fed into GAT \cite{velickovic2018graph} to obtain frame-level graph features $\mathrm{F}_{\mathrm{G_{f}}} =\{\mathrm{F}_{\mathrm{G_{f}^1}},\dots,\mathrm{F}_{\mathrm{G_{f}^N}}\}$. We introduce to apply transformer \cite{vaswani2017attention} on $\mathrm{F}_{\mathrm{G_{f}}}$ to further include the temporal dependency between frame-level graph features and give the final representation $\mathbf{R}_{\mathrm{G_{f}}}$ for frame-level graph $\mathrm{G_{f}}$.

We also construct a holistic video-level graph $\mathrm{G_{v}}$ to capture fine-grained temporal node connections, which builds edges between nodes is not only one single frame but also adjacent frames. Specifically, we compute the cosine similarity and Interaction over Union (IoU) between node pairs from adjacent frames. If the computed similarity and IoU are greater than the pre-defined thresholds, we connect them together. By this way, we group all $\mathrm{G_{f}}$ together and build $\mathrm{G_{v}}$. We also adopt GAT \cite{velickovic2018graph} to encode $\mathrm{G_{v}}$ and give $\mathbf{R}_{\mathrm{G_{v}}}$. The output representation of this module $\mathbf{R_{obj}}$ is the concatenation of $\mathbf{R}_{\mathrm{G_{f}}}$ and $\mathbf{R}_{\mathrm{G_{v}}}$.

\section{Experiments}
We evaluate the efficacy of our proposed framework in two public datasets: MSR-Video To Text (MSR-VTT) dataset \cite{xu2016msr} and Microsoft Video Description (MSVD) dataset \cite{chen2011collecting}. The results are obtained from four captioning metrics: BLEU, METEOR, ROUGE-L and CIDEr. “-” means number is not available. The reported results are evaluated with the Microsoft COCO evaluation server \cite{chen2015microsoft}.
We compare our results with the previous state-of-the-art models and report extensive ablation studies to show the effectiveness of each module of our model. 

\subsection{Datasets}
\noindent \textbf{MSR-VTT} \cite{xu2016msr}. MSR-VTT is a commonly used benchmark dataset for video captioning task. It is composed of $10,000$ video clips, where each video clip is annotated with $20$ English text. These video clips are categorized into $20$ classes, such as music, cooking and etc. We follow the standard splits \cite{wang2019controllable,zhang2020object,zheng2020syntax,pan2020spatio}, i.e. there are $6,513$, $497$ and $2,990$ for training, validation and testing respectively.  

\noindent \textbf{MSVD} \cite{chen2011collecting}. MSVD is a relatively small-scale dataset compared with MSR-VTT, as it in total contains $1,970$ video clips. MSVD has multilingual captions, while we only consider the English annotations. 
There are roughly $40$ English sentences for each video clip. 
Similar with prior work \cite{wang2019controllable,zhang2020object,zheng2020syntax,pan2020spatio}, the dataset is separated into $1,200$ training clips, $100$ validation clips and $670$ test clips.

\begin{table*}[htbp!]
  \centering
    \caption{\textbf{Main results.} Evaluation of performance compared against various baseline models on the MSR-VTT dataset, we evaluate the results with BLEU@1$\sim$4, METEOR, ROUGE-L and CIDEr scores (\%). We also state the video context features used by the listed methods, where V, G, R-\textit{N}, A, Ca, IRV2 and RoI denote VGG19, GoogleNet, \textit{N}-layer ResNet, Audio, Category, InceptionResnetV2 and Region of Interest (RoI) features respectively. BERT and h-LSTMs denote BERT pretrained model and the hierarchical-LSTMs respectively. XE and RL denote training with cross-entropy loss and reinforcement learning respectively.}
  \begin{center}
  \scalebox{1.0}{
  \begin{tabular}{l|ccccccccc}
  \toprule
\textbf{Model} & \textbf{Backbone} & \textbf{BLEU@1}  & \textbf{BLEU@2}  & \textbf{BLEU@3}   & \textbf{BLEU@4}  & \textbf{Meteor} &  \textbf{Rouge-L} & \textbf{CIDEr}    & \textbf{Training} \\ \hline \hline
SA \cite{yao2015describing} & V+C3D           & 72.2 & 58.9 & 46.8  & 35.9 & 24.9  & -    & -    & XE \\
M3 \cite{wang2018m3}  & V+C3D           & 73.6 & 59.3 & 48.26 & 38.1 & 26.6 & -    & -    & XE \\
MA-LSTM \cite{xu2017learning} & G+C3D+A         & -    & -    & -     & 36.5 & 26.5  & 59.8 & 41.0 & XE \\
VideoLab \cite{ramanishka2016multimodal}  & R-152+C3D+A+Ca  & -    & -    & -     & 39.1 & 27.7  & 60.6 & 44.1 & XE \\
v2t\_navigator \cite{jin2016describing} & C3D+A+Ca        & -    & -    & -     & 42.6 & 28.8  & 61.7 & 46.7 & XE \\
RecNet \cite{wang2018reconstruction} & InceptionV4     & -    & -    & -     & 39.1 & 26.6  & 59.3 & 42.7 & XE \\
OA-BTG \cite{zhang2019object} & R-200+RoI       & -    & -    & -     & 41.4 & 28.2  & -    & 46.9 & XE \\
MARN \cite{pei2019memory} & R-101+C3D+Ca    & -    & -    & -     & 40.4 & 28.1  & 60.7 & 47.1 & XE \\
MGSA \cite{chen2019motion}  & IRV2+C3D        & -    & -    & -     & 42.4 & 27.6  & -    & 47.5 & XE \\
STG \cite{pan2020spatio} & IRV2+I3D+RoI   & -    & -    & -     & 40.5 & 28.3  & 60.9 & 47.1 & XE \\
ORG-TRL \cite{zhang2020object}   & IRV2+C3D+RoI+BERT    & -    & -    & -     & 43.6 & 28.8  & 62.1 & 50.9 & XE \\
POS-CG  \cite{wang2019controllable}  & IRV2+I3D+Ca     & 79.1 & 66.0 & 53.3  & 42.0 & 28.1  & 61.1 & 49.0 & XE \\
SAAT \cite{zheng2020syntax}  & IRV2+C3D+Ca+RoI & 80.2 & 66.2 & 52.6  & 40.5 & 28.2  & 60.9 & 49.1 & XE \\
SibNet \cite{liu2020sibnet} & Temporal networks & -    & -    & -  & 41.2 & 27.8  &  60.8 & 48.6 & XE \\
RCG \cite{zhang2021open} & IRV2+C3D+h-LSTMs & -    & -    & -  & 43.1 & 29.0  &  61.9 & 52.3 & XE \\
CMG (ours) & IRV2+C3D & 79.2 & 65.7 & 52.6 & 40.9 & 28.7 & 61.3 & 49.2 & XE \\
CMG (ours) & IRV2+C3D+Ca & 81.6 & 67.0 & 54.4 & 43.1 & 29.2 & 61.8 & 51.5 & XE \\
CMG (ours) & IRV2+C3D+A+Ca & \textbf{83.5} & \textbf{70.7} & \textbf{57.4} & \textbf{44.9} & \textbf{29.6} & \textbf{62.9} & 53.0 & XE  \\
\midrule
PickNet \cite{chen2018less} & R-152+Ca  & -    & -    & -    & 41.3 & 27.7 & 59.8 & 44.1 & RL \\
SAAT  \cite{zheng2020syntax} & IRV2+C3D+Ca+RoI & 79.6 & 66.2 & 52.1 & 39.9 & 27.7 & 61.2 & 51.0 & RL \\
POS-CG \cite{wang2019controllable} & IRV2+I3D+Ca  & 81.2 & 67.9 & 53.8 & 41.3 & 28.7 & 62.1 & 53.4 & RL \\
CMG (ours) &  IRV2+C3D+A+Ca  & 83.4 & 70.1 & 56.3 & 43.7 & 29.4 & 62.8 & \textbf{55.9} & RL \\
  \bottomrule
  \end{tabular}
  }
  \end{center}
  \label{table:main}
\end{table*}

\subsection{Implementation Details}
\subsubsection{Feature Extraction}
We follow \cite{phan2017consensus,zheng2020syntax} to extract video context features for MSR-VTT dataset, and use four types of features. Specifically, we extract 2D features from the last avg-pooling layer of pretrained InceptionResnetV2 (IRV2) \cite{szegedy2016inception}. We adopt a C3D \cite{tran2015learning} model pretrained on Sports-1M \cite{KarpathyCVPR14} dataset to capture short-term motion features. In terms of audio features, they are extracted from audio segments within frame steps from MFCC \cite{davis1980comparison}. Since MSR-VTT provides category information, we also use GloVe \cite{pennington2014glove} to encode the semantic labels for each video. 

For MSVD, we take 2D and 3D visual features as the video context features following the previous practice \cite{zheng2020syntax,chen2019semantics}. Since it has limited number of training video clips, we only take two features to avoid over-fitting in this dataset, i.e. ResNeXt \cite{xie2017aggregated} pretrained on the ImageNet dataset is adopted to extract visual features, an ECO \cite{zolfaghari2018eco} pretrained on the Kinetics400 dataset is used to give video temporal features. Specifically, we use $32$ evenly extracted video frames as the input, which are fed into ResNeXt and ECO respectively. We take the averaged ResNeXt conv5/block3 output as 2D visual features and the global pool results of ECO as 3D features. 

\subsubsection{Model Setting} 
We pick $N=10$ keyframes to be the input based on the difference between frames, and we take top $K=60$ synonym categories out of $\mathrm{T}$ to be semantic classes of cross-modal meta concepts. 

In the weakly-supervised meta concept learning, we train the model with batch size of $60$ and learning rate of $4\times10^{-4}$, and set the parameter $m$ and $\lambda$ as $0.3$ and $0.5$ respectively. Specifically, we use the output feature maps from the ResNet-101 last convolutional layer as the video frame feature representations, the final spatial dimension is $14 \times 14$ and the feature dimension is $2048$. Then we randomly sample $4$ frame features and input these into the weakly-supervised meta concept learning model. The model aims to generate captions and produce the attended regions $\alpha_i$ with the size of $(4, 14, 14)$ for each caption token. 
We set a threshold to alleviate the noise in $\alpha$ heatmap. To specific, the threshold is set as $80$, the heatmap values smaller than $80$ are set as background. When we use $\alpha$ as the pseudo masks for training, we resize them to the same size as the input video frames.
We train semantic segmentation model PSPNet \cite{zhao2017pyramid} to localize the learned meta concepts. The segmentation model is trained with batch size of $8$ and learning rate of $0.05$. 

In the dynamic meta concept graph construction, we allow each node to connect its $J=3$ nearest neighbour and output $256$-dimensional features. The adopted scene graph detector \cite{tang2020sggcode,tang2020unbiased} is pretrained on Visual Genome (VG) dataset \cite{krishna2017visual} with Faster R-CNN \cite{ren2015faster} backbone. We use a two-layer graph attention networks (GAT) \cite{velickovic2018graph} to encode the constructed video graphs, where we set the hidden dimension as $8$, head number as $8$ and output dimension as $256$. Then we use a one-layer transformer \cite{vaswani2017attention} with $4$ heads to give temporal representations of frame-level graphs.

In the caption decoder, we use word embedding layer to give word representations, whose dimension is $512$. We also map all the used visual context features onto the space of $512$-dimensional space and then concatenate them together to be decoder input. We take a one-layer plain LSTM as the decoder. 
We train the decoder with batch size of $32$ and learning rate of $8\times10^{-5}$.
Our implemented reinforcement learning (RL) strategy is based on SCST \cite{luo2018discriminability}. We use beam search for evaluation, and set the beam size as $5$. 

\subsubsection{Model Efficiency} 
We run all the experiments on a single V100 GPU. We need to train three components in our proposed framework.
For the weakly-supervised meta concept learning module, we set the training epoch number as $120$. The total training process costs about $3$ hours. 
For the meta concept localization (segmentation) model, we set the training epoch number as $80$. The total training process costs about $1$ day.
For the captioning model training, we run $50$ epochs for the non-RL and RL settings respectively. In the non-RL training, the total training process costs about $2$ hours. In the RL training, the total training process costs about $8$ hours. During inference phase, each sample costs about $0.1$ s.
To summarize, our proposed method is efficient for training and inference.


\begin{table}
\caption{Performance comparisons with different baseline methods on the testing set of the MSVD dataset. The results are evaluated with BLEU@4, METEOR, ROUGE-L and CIDEr scores (\%).}
  \label{table:msvd}
  \begin{center}
  \scalebox{1.1}{
  \begin{tabular}{l|cccc}
  \toprule
  \textbf{Model}& \textbf{B@4} &\textbf{M} & \textbf{R} &  \textbf{C}  \\ 
  \midrule
    MA-LSTM \cite{xu2017learning}  &52.3  &33.6  &-  &70.4  \\
    MGSA \cite{chen2019motion}  & 53.4 & 35.0 & - & 86.7\\
    OA-BTG \cite{zhang2019object} & 56.9 & 36.2 & - & 90.6 \\
    POS-CG  \cite{wang2019controllable} &52.5  &34.1  &71.3  &88.7  \\
    SAAT \cite{zheng2020syntax} &46.5 & 33.5 & 69.4 & 81.0  \\
    STG \cite{pan2020spatio} &52.2 & 36.9 & 73.9 & 93.0 \\ 
    ORG-TRL \cite{zhang2020object} &54.3 & 36.4 & 73.9 & 95.2 \\
    SibNet \cite{liu2020sibnet} & 55.7 & 35.5 & 72.6 & 88.8 \\
    \hline
    CMG (ours) & \textbf{59.5} & \textbf{38.8}  & \textbf{76.2} & \textbf{107.3}  \\
    \bottomrule
  \end{tabular}
  }
  \end{center}
\end{table}

\subsection{Experimental Results}
\subsubsection{Performance Comparison} 
In Table \ref{table:main} we compare our results against earlier models under different training strategies, i.e. cross-entropy loss and reinforcement learning, on MSR-VTT dataset \cite{xu2016msr}. In table \ref{table:msvd}, we show model performance in MSVD dataset \cite{chen2011collecting}. It can be observed that our results gain remarkable improvement across various metrics on both MSR-VTT and MSVD datasets. We also conduct experiments with different video backbone features, to indicate the importance of multi-sourced information.  

STG \cite{pan2020spatio} and ORG-TRL \cite{zhang2020object} models utilize similar methodologies, which construct graphs based on object proposals. STG uses a transformer \cite{vaswani2017attention} as the decoder, while ORG-TRL uses an extra pretrained BERT \cite{devlin2018bert} model, it gets higher results than STG. Another reason for STG having relatively low performance is that, MSR-VTT contains a large portion of animations, making pretrained object detection models often fail in these scenarios. When we use similar video context features (\emph{IRV2+C3D}) as STG, our model outperforms STG by around $5 \%$ in CIDEr score, indicating our proposed cross-modal meta concepts can be adapted to different datasets and help alleviate such issues even without external language model.

To enable the generation model to keep aware of the syntax information, Zheng et al. \cite{zheng2020syntax} adopt predicate and object information to guide the language decoder in generation. In SAAT \cite{zheng2020syntax}, instead of construing graph representations, they directly use the attention mechanism to encode the predicted predicates and objects as the input vectors for decoder. In contrast, when we use similar context features (\emph{IRV2+C3D+Ca}) as SAAT, our model can outperform SAAT by a margin, where we propose to build holistic and local video graphs for the predicted syntax, indicating the effectiveness of our model. RCG \cite{zhang2021open} incorporates the retrieval learning into the captioning process, where they use \emph{IRV2+C3D} as the video context features and adopt the hierarchical-LSTMs to generate captions. While here we only use the plain LSTM for the caption generation, in order to make fair comparisons with most of the previous works \cite{zheng2020syntax,wang2019controllable,chen2019motion,pei2019memory}. It may indicate better decoder can give better generation results.

When we shift to the RL training strategy that directly optimizes our model with CIDEr scores, we achieve the highest CIDEr score. In general, the performance of our proposed model is shown to be very promising, having improvements in all metrics consistently.

\begin{table}
  \centering
  \caption{\textbf{Ablation Studies.} Evaluation of the benefits of different modules of the proposed model, where $J$, MC, FG and VG denote the number of connected neighbours, meta concept graphs, frame-level and video-level graphs respectively.
  We also show the results of MC with and without the dynamic graph encoding or visual/semantic features.
  The results are evaluated with BLEU@4, METEOR, ROUGE-L and CIDEr scores (\%) on MSR-VTT dataset.
  }
  \label{table:ablation}
  \begin{center}
  \scalebox{1.1}{
  \begin{tabular}{l|cccc}
  \toprule
\textbf{Method}  & \textbf{B@4} & \textbf{M} & \textbf{R} & \textbf{C} \\
\midrule
Baseline (BL) & 43.0  & 28.1 & 61.5  & 50.2  \\
\hline
BL + MC ($J=20$) &  44.0 & 29.5 & 62.5 & 51.9 \\
BL + MC ($J=10$) &  44.1 & 29.5 & 62.6 & 52.0  \\
BL + MC ($J=3$)  & 44.7 & 29.4 & 62.9 & 52.2 \\
\quad \quad \quad - attention-LSTM & 43.6 & 29.1 & 62.4 & 51.3 \\
\quad \quad \quad - Semantic Only & 43.1 & 28.9 & 62.0 & 51.0 \\
\quad \quad \quad - Visual Only & 43.7 & 29.1 & 62.4 & 51.6 \\
\hline
BL + FG  & 43.8 & 29.1 & 62.1 & 51.4 \\
BL + VG (no rel) & 43.3 & 29.0  & 61.9 & 51.2 \\
BL + VG  & 44.0 & 29.1  & 62.2 & 51.7 \\
BL + FG + VG & 44.3 & 29.1 & 62.4 & 51.8 \\
\hline
All   & \textbf{44.9} & \textbf{29.6} & \textbf{62.9}  & \textbf{53.0} \\
  \bottomrule
  \end{tabular}
  }
  \end{center}
\end{table}

\begin{figure*}[htbp!]
\begin{center}
\includegraphics[width=0.7\textwidth]{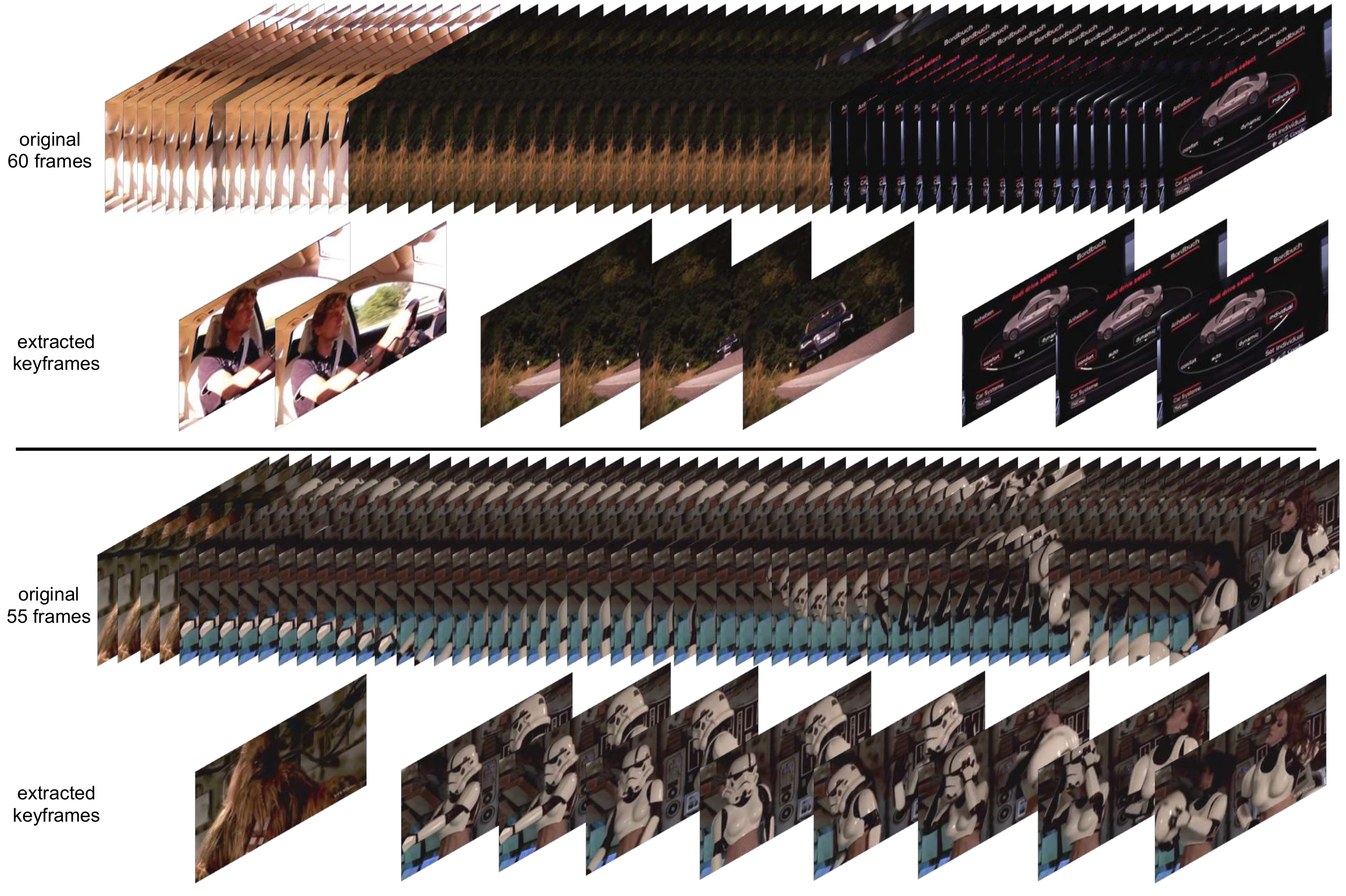}
\vspace{-0.1in}
\end{center}
  \caption{\textbf{Qualitative analysis of the selected keyframes.} The upper rows show the original full video frames, the bottom rows show our generated keyframes.}

\label{fig:keyframe}
\end{figure*}

\begin{table}
  \centering
  \caption{Caption generation performance of weakly-supervised meta concept learning evaluated with BLEU@4 and Top-5 accuracy (\%) at MSR-VTT and MSVD datasets, where CA denotes cross-modal alignment.
  }
  \label{table:weakly}
  \begin{center}
  \scalebox{1.1}{
  \begin{tabular}{lc|ccc}
  \toprule
& \textbf{Methods}  & \textbf{B@4} & \textbf{Top-5 Acc} \\
\midrule
\multirow{2}{*}{MSR-VTT} & without CA & 15.7 & 61.5          \\
                         & with CA   & \textbf{16.3} & \textbf{63.3}         \\
\midrule
\multirow{2}{*}{MSVD}    & without CA & 24.8  & 68.6        \\
                         & with CA  & \textbf{26.1} & \textbf{70.7}         \\
  \bottomrule
  \end{tabular}
  }
  \end{center}
\end{table}

\subsubsection{Ablation Studies}
We conduct extensive ablation studies as shown in Table \ref{table:ablation} and \ref{table:weakly} and \ref{table:ablation_l}.

\noindent \textbf{Effectiveness of the meta concept graph.} To observe the impact of the number of neighbours $J$ in the dynamic meta concept (MC) graph embedding process, we change $J$ to different values. The results show there is minor difference between different settings for $J$, one possible reason is: with adaptive updating on node embeddings and edge connections, nodes can aggregate around semantically similar node features. To validate the efficacy of our dynamic graph construction method, we follow \cite{zhou2019grounded} to build an attention-LSTM to encode the learned cross-modal meta concepts, denoting as \emph{- attention-LSTM} in Table \ref{table:ablation}, for comparison. 
It can be seen that the dynamic graph encoding method of our framework gives better performance than the attention-LSTM embedding method, indicating the graph embedding gives better feature aggregation.  
Besides, we also evaluate the usefulness of our learned visual and semantic concepts separately, the results suggest that visual features can give better performance than pure semantic features.

\noindent \textbf{Effectiveness of scene graph.} We compare the performance of video-level graphs (VG) without and with predicate information, which corresponds to VG (no rel) and VG respectively. In VG (no rel), which is similar with the setting in \cite{pan2020spatio,zhang2020object}, we only connect object nodes together without using predicate relationships. We observe that with predicates, our model can gain better results.    
We incrementally add frame-level graphs (FG) and VG on the baseline (BL) respectively, which can be seen both types of graphs boost baseline performance. VG give better scores regarding BLEU@4 and CIDEr, showing that graphs with more informative connections can output better representations.
On the whole, we can see that each proposed module gives positive effects to our model by improving captioning performance, and they can work collaboratively with other modules to output overall boosted results.

\begin{figure*}
\begin{center}
\includegraphics[width=0.7\textwidth]{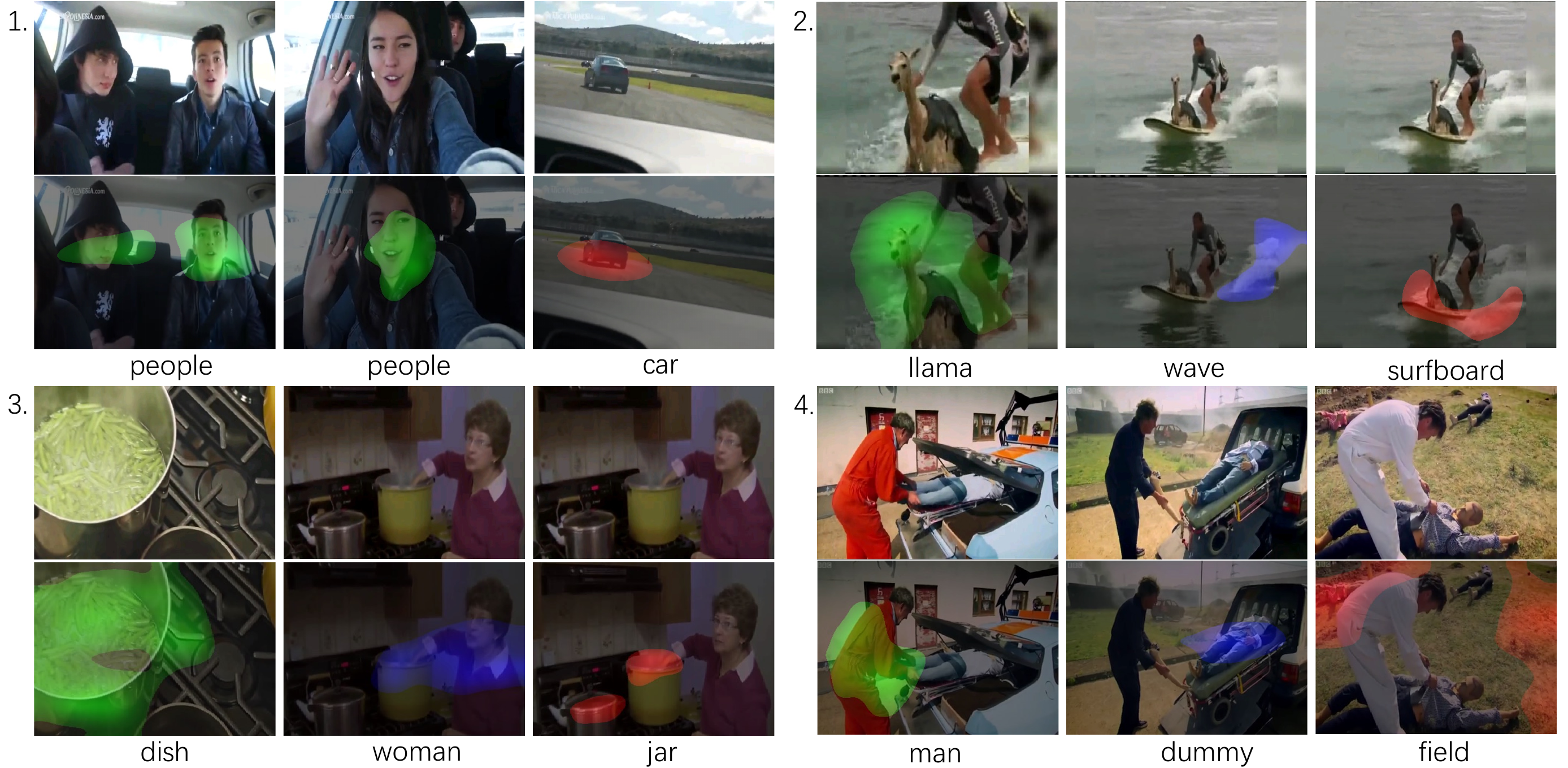}
\end{center}
  \caption{\textbf{Qualitative analysis of the segmented regions of our proposed weakly-supervised learning method.} We present 4 groups of video frames and the corresponding segmented visual regions of the given semantic meta concepts. The first row are the picked video keyframes and the second row stated the learned visual meta concepts, which are the visual segmented areas. The segmented areas are obtained through the weakly-supervised learning method, which are used as the meta concepts in our proposed CMG video captioning framework.}

\label{fig:region_vis}
\end{figure*}

\begin{table}
  \centering
  \caption{Ablation study on $\lambda$ of weakly-supervised meta concept learning model. The performance is evaluated with BLEU@4 and Top-5 accuracy (\%) at the MSR-VTT dataset.
  }
  \label{table:ablation_l}
  \begin{center}
  \scalebox{1}{
  \begin{tabular}{c|ccc}
  \toprule
 \textbf{$\lambda$}  & \textbf{B@4} & \textbf{Top-5 Acc} \\
\midrule
  0.1 & 15.7 &    61.9   \\
 0.5 & \textbf{16.3} & \textbf{63.3}         \\
 1.0  & 16.0 &   62.4     \\
  \bottomrule
  \end{tabular}
  }
  \end{center}
  \vspace{-0.2in}
\end{table}

\noindent \textbf{Evaluation of weakly-supervised meta concept learning.} In Table \ref{table:weakly}, we show the efficacy of our proposed cross-modal alignment (CA) for weakly-supervised meta concept learning. In Table \ref{table:ablation_l}, we show the ablation study on $\lambda$. It is hard to evaluate the quality for learned cross-modal meta concepts directly, as we do not have any ground truth. Hence we choose to evaluate the model caption generation performance, since the sequence is produced based on localized meta concepts, such that the generation results can reflect the quality of learned cross-modal meta concepts to some extent. 
It can be observed that CA can help improve captioning performance on both MSR-VTT and MSVD datasets, and setting $\lambda=0.5$ gives the best results. 

\begin{figure*}[htbp!]
\begin{center}
\includegraphics[width=0.7\textwidth]{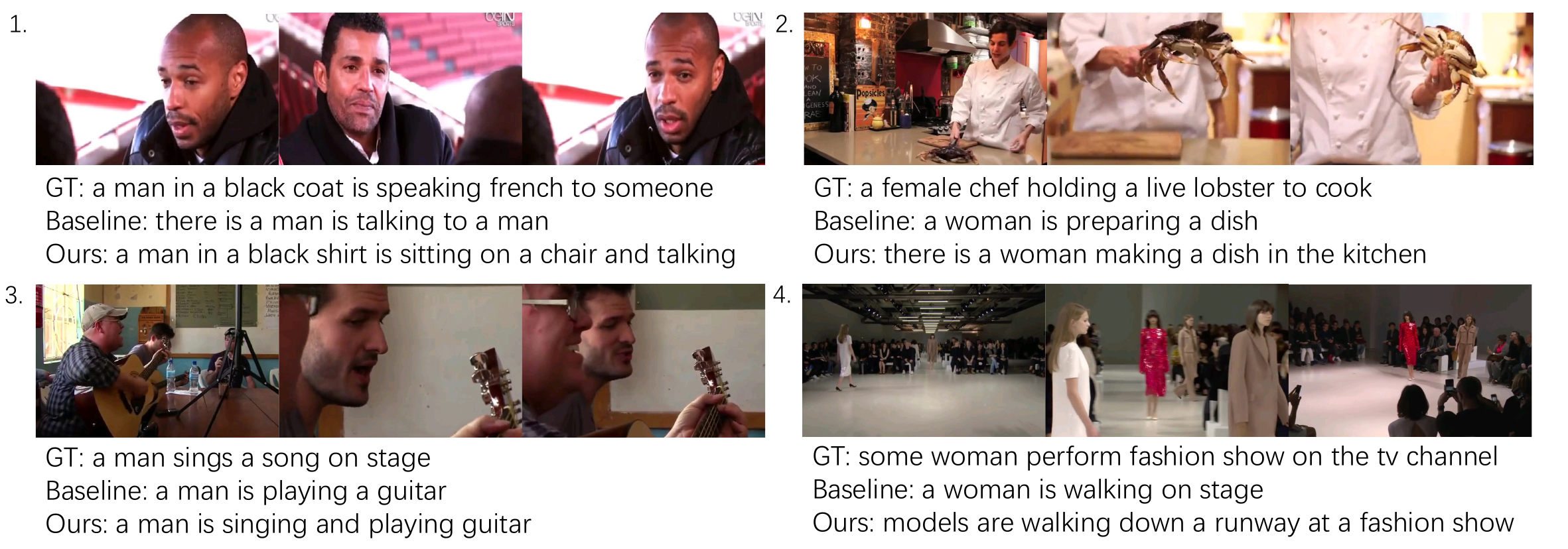}
\end{center}
  \caption{\textbf{Qualitative analysis of our generated video captions.} We present 4 groups of video frames and their ground truth (GT), baseline and our model generated video captions. For the baseline model, we do not apply the proposed cross-modal graph framework on it, and it produces relatively short captions, which may lose some semantic attributes.}
\label{fig:cap_vis}
\end{figure*}

\begin{figure}
\begin{center}
\includegraphics[width=0.35\textwidth]{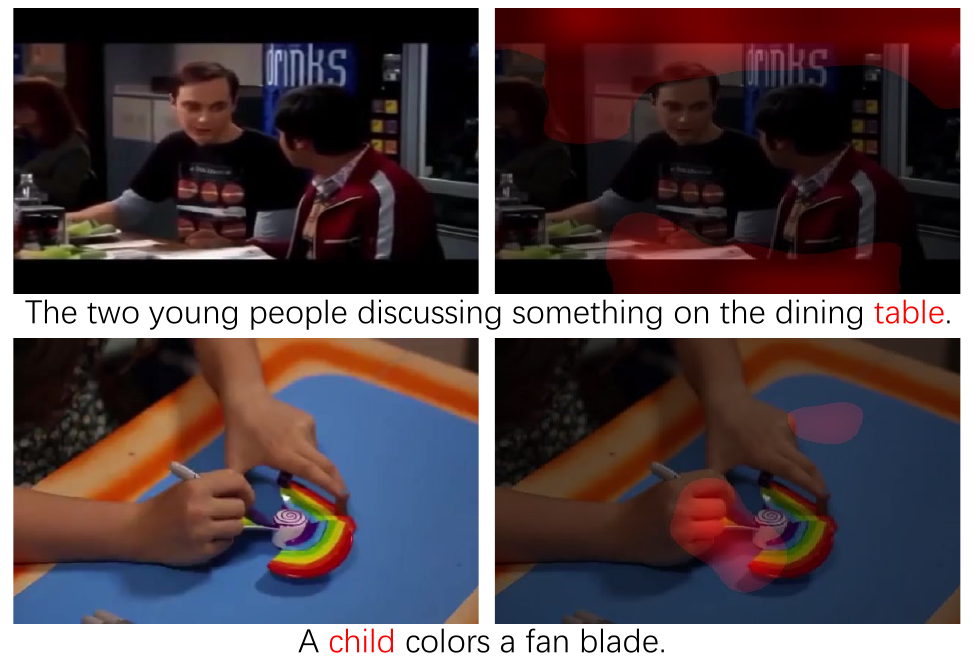}
\end{center}
  \caption{\textbf{Qualitative analysis of failure cases.} We indicate the cross-modal meta concepts with the same colour. It is observed our model can hardly give reasonable attended visual regions for items that are not shown in the video frames.}
\label{fig:fail}
\end{figure}

\begin{figure*}[htbp!]
\begin{center}
\includegraphics[width=0.8\textwidth]{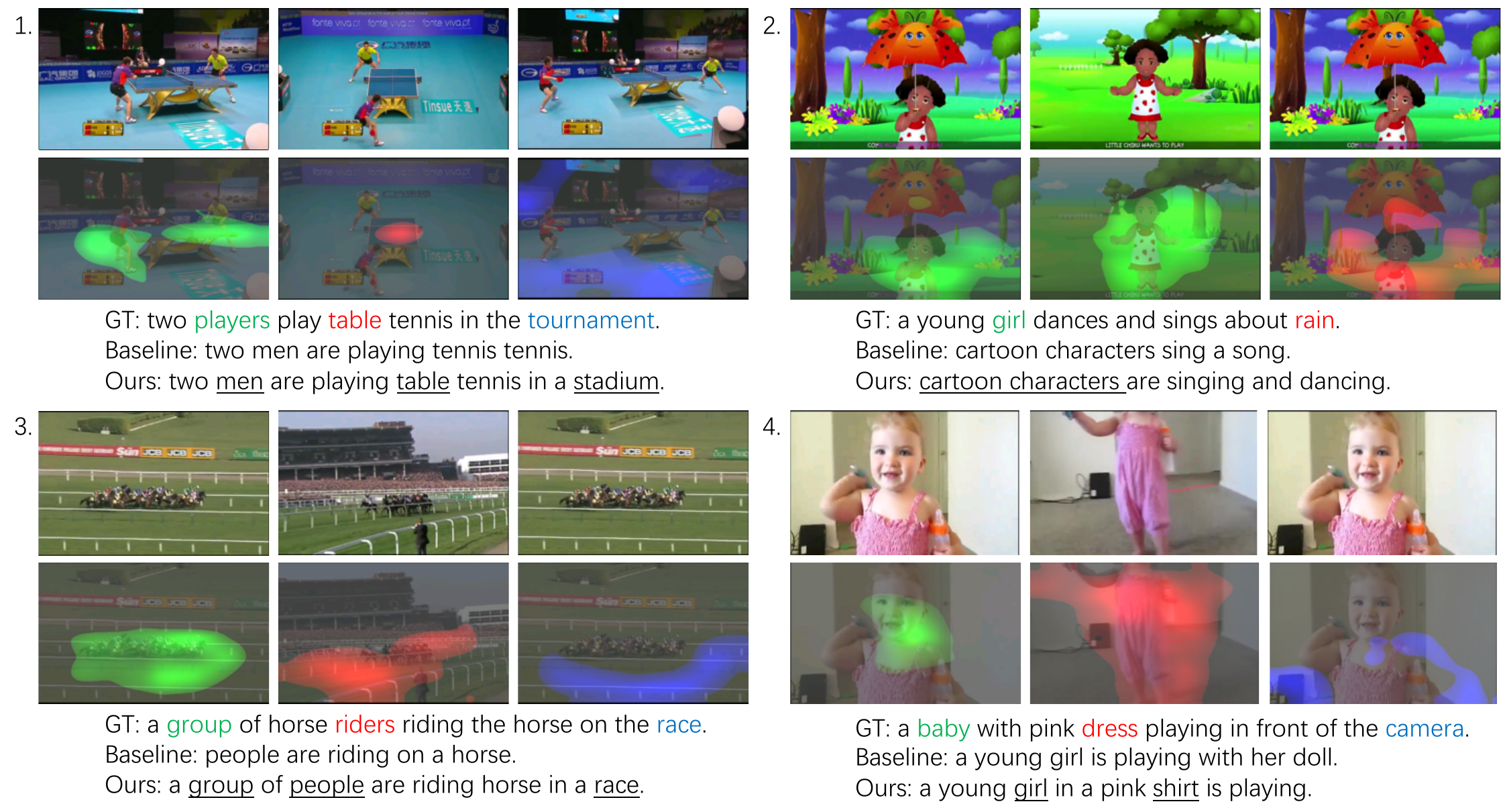}
\end{center}
  \caption{\textbf{Qualitative analysis of the learned cross-modal meta concepts and our generated captions.} We present 4 groups of video frames and their ground truth (GT), baseline and our model generated captions, where the first row are the picked keyframes and the second row stated the learned visual meta concepts. 
  We indicate the cross-modal meta concepts with the same color in the GT captions, and the underline generated words denote correspondence with the localized meta concepts.}
\label{fig:vis}
\end{figure*}

\subsubsection{Qualitative Results}
We show the qualitative results in Figure \ref{fig:keyframe}, \ref{fig:region_vis}, \ref{fig:cap_vis}, \ref{fig:fail} and \ref{fig:vis}.

\noindent\textbf{The demonstration of our generated keyframes.} In Figure \ref{fig:keyframe}, we present qualitative results to demonstrate the efficacy of our keyframe extraction method. Specifically, we decoded videos with 5 FPS. According to our observations, the decoded video frames have repeated information. Our difference-based keyframe extraction method appears to successfully identify the change of actions and scenes. For example, in the upper row of Figure \ref{fig:keyframe}, the extracted keyframes show three different scenes of the given video, i.e., 1) a man is driving a car, 2) a car is running on the road, and 3) a demonstration of the car.

\noindent \textbf{The visualization of the weakly-learned visual meta concepts.} In Figure \ref{fig:region_vis}, we present the learned visual meta concepts with the weakly-supervised learning method. Since the only supervision for the attended semantic visual regions is the given video captions, we do not expect precise pixel labelling for the regions. However, we can still observe the proposed framework outputs mostly reasonable results. To be specific, we show the most activated regions across the video frames of the given semantic meta concepts. For example, in the third sample, the model gives the coarse region of the given semantic meta concept \emph{dish}, which is not defined by the prevailing object detectors. Besides, based on the given captions, the model can also localize the visual regions of \emph{jar}. These predicted visual meta concepts help give more fine-grained information compared to the traditional object detectors, which allows the model to learn different classes of meta concepts for different datasets. This property enables our proposed framework to produce some semantic information that is missed by the existing object detection methods.

\noindent \textbf{The visualization of the generated captions from the videos.} In Figure \ref{fig:cap_vis}, we present the visualization examples of our generated video captions. For each video, we show the picked keyframes, ground truth (GT) captions and the captions generated by the baseline model and our proposed CMG model respectively. To be specific, in the first example, our generated caption also gives the information about the person's clothes, which is the \emph{black shirt}, while the baseline results lose such fine-grained attributes. In the last example, our generated captions produce more descriptions on the video context: \emph{a fashion show}. We observe our generated video captions generally contain more useful semantic information than the baseline results. The proposed cross-modal graph with the learned meta concepts guides the video captioning model to focus on the fine-grained semantic attributes of the video frames, hence it allows the model to produce textual descriptions with more sufficient desired information than the baseline model.

\noindent \textbf{The visualization of the correspondence between the generated visual and semantic meta concepts.} In Figure \ref{fig:vis}, we show the qualitative analysis for our learned visual and semantic meta concepts, where we visualize the video frames from different scenarios. Specifically, in the second example that is an animation clip, our learned cross-modal meta concepts can localize the visual regions of cartoon characters, while some pretrained object detection model may fail \cite{pan2020spatio}. The learned meta concepts also allow the generation model to keep aware of the visual context information, such as \emph{tournament} in the first video and \emph{race} in the third one, which generate precise captioning words on the video context. 
Generally, the learned cross-modal meta concepts show promising results, where the CMG model gives more useful textual descriptions than the baseline model, thus boosts the model captioning performance.

\noindent \textbf{Failure cases.} Figure \ref{fig:fail} and the last example of Figure \ref{fig:vis} are failure cases. We observe the proposed model can hardly give reasonable attended visual regions for items that are not shown in the video frames, for instance, the \emph{table} in the upper row of Figure \ref{fig:fail}. However, there are only small amounts of failure cases, the learned meta concepts are demonstrated to improve the baseline model performance by around 4\% at CIDEr score.


\section{Conclusion}
In this paper, we propose CMG with meta concepts for video captioning. Specifically, we use a weakly-supervised learning approach to localize the attended visual regions and their semantic classes for objects shown in captions, in an attempt to cover some undefined classes of pretrained models. We then use dynamic graph embeddings to aggregate semantically similar nodes and give meta concept representations. To include predicate relationships between objects, we adopt detected scene graphs in frames to build video- and frame-level graphs and give structure representations. We conduct extensive experiments and ablation studies, and achieve state-of-the-art results on MSR-VTT and MSVD datasets for video captioning.

\section*{Acknowledgments}
This research is supported, in part, by the National Research Foundation (NRF), Singapore under its AI Singapore Programme (AISG Award No: AISG-GC-2019-003) and under its NRF Investigatorship Programme (NRFI Award No. NRF-NRFI05-2019-0002). Any opinions, findings and conclusions or recommendations expressed in this material are those of the authors and do not reflect the views of National Research Foundation, Singapore. This research is supported, in part, by the Singapore Ministry of Health under its National Innovation Challenge on Active and Confident Ageing (NIC Project No. MOH/NIC/HAIG03/2017).
This research is also supported by the National Research Foundation, Singapore under its AI Singapore Programme (AISG Award No: AISG-RP-2018-003), and the MOE AcRF Tier-1 research grant: RG95/20.

\bibliographystyle{IEEEtran}
\bibliography{egbib}


\vspace{11pt}

\begin{IEEEbiography}[{\includegraphics[width=1in,height=1.25in,clip,keepaspectratio]{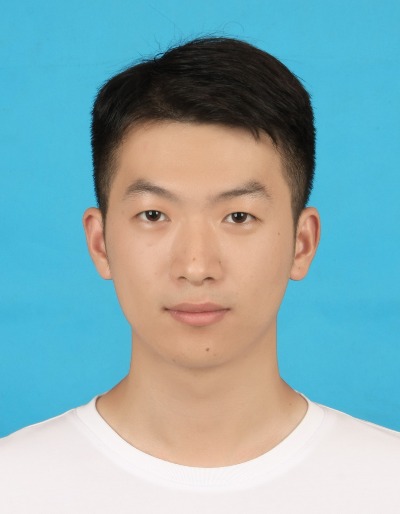}}]{Hao Wang}
is a PhD candidate with the School of Computer Science and Engineering, Nanyang Technological University, Singapore. His research interests include cross-modal generation and computer vision.
\end{IEEEbiography}

\begin{IEEEbiography}[{\includegraphics[width=1in,height=1.25in,clip,keepaspectratio]{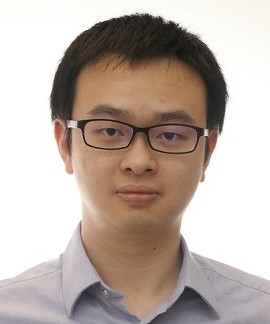}}]{Guosheng Lin} 
is currently an Assistant Professor with the School of Computer Science and Engineering, Nanyang Technological University, Singapore. His research interests include computer vision and machine learning.
\end{IEEEbiography}

\begin{IEEEbiography}[{\includegraphics[width=1in,height=1.25in,clip,keepaspectratio]{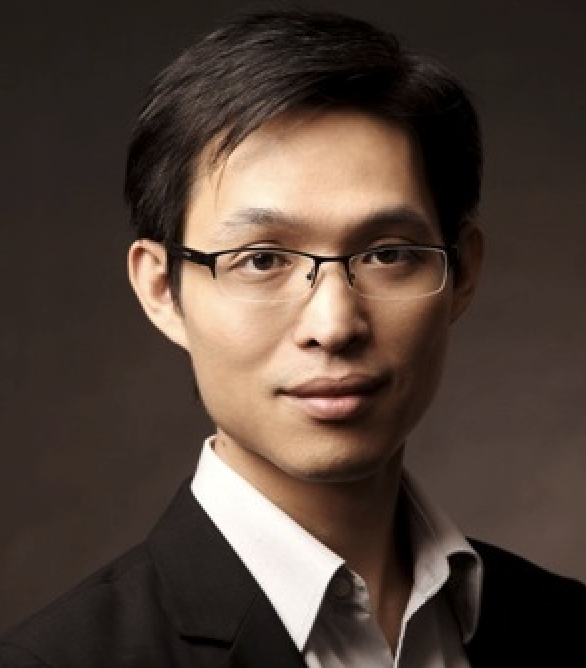}}]{Steven C. H. Hoi}
is currently the Managing Director of Salesforce Research Asia, and a Professor of Information Systems at Singapore Management University, Singapore. He has served as the Editor-in-Chief for Neurocomputing Journal, guest editor for ACM Transactions on Intelligent Systems and Technology. He is an IEEE Fellow and ACM Distinguished Member.
\end{IEEEbiography}

\begin{IEEEbiography}[{\includegraphics[width=1in,height=1.25in,clip,keepaspectratio]{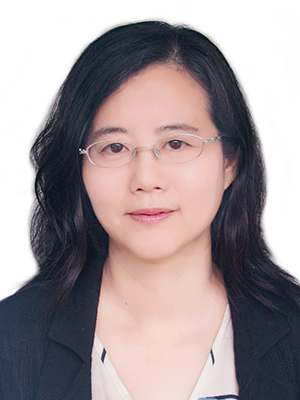}}]{Chunyan Miao} 
is the chair of School of Computer Science and Engineering in Nanyang Technological University (NTU), Singapore. Dr. Miao is Director of the Joint NTU-UBC Research Centre of Excellence in Active Living for the Elderly (LILY), Nanyang Technological University (NTU), Singapore. She is the Editor-in-Chief of the International Journal of Information Technology published by the Singapore Computer Society.
\end{IEEEbiography}

\vfill

\end{document}